\documentclass{article}
\usepackage{PRIMEarxiv}

\usepackage[utf8]{inputenc}
\usepackage[T1]{fontenc}

\usepackage{amsmath,amsfonts}

\usepackage{graphicx}
\usepackage{array}
\usepackage{multirow}
\usepackage[caption=false,font=normalsize,labelfont=sf,textfont=sf]{subfig} 

\usepackage{textcomp}
\usepackage{verbatim}
\usepackage{float}
\usepackage{paralist}
\usepackage[inline]{enumitem}
\usepackage{balance}
\usepackage{siunitx}
\usepackage{booktabs} 
\usepackage[compress]{cite}
\usepackage{url}           

\usepackage{algorithmic}

\usepackage{xcolor}
\usepackage[hidelinks]{hyperref}
\usepackage{orcidlink}

\title{Explainability-Driven Dimensionality Reduction for Hyperspectral Imaging}

\author{
  Salma Haidar \orcidlink{0000-0003-4578-8877}\thanks{Corresponding author: salma.haidar@uantwerpen.be}
  \qquad
  Jos{\'e} Oramas \orcidlink{0000-0002-8607-5067}\\
  Department of Computer Science\\
    University of Antwerp, imec - IDLab \\
    Sint-Pietersvliet 7,2000, Belgium\\	
  {\tt\small \{salma.haidar,jose.oramas\}@uantwerpen.be}
}

\begin{document}
\maketitle

\begin{abstract}
 
 Hyperspectral imaging (HSI) provides rich spectral information for precise material classification and analysis; however, its high dimensionality introduces a computational burden and redundancy, making dimensionality reduction essential. We present an exploratory study into the application of post-hoc explainability methods in a model-driven framework for band selection, which reduces the spectral dimension while preserving predictive performance. A trained classifier is probed with explanations to quantify each band’s contribution to its decisions. We then perform deletion–insertion evaluations, recording confidence changes as ranked bands are removed or reintroduced, and aggregate these signals into influence scores. Selecting the highest-influence bands yields compact spectral subsets that maintain accuracy and improve efficiency. Experiments on two public benchmarks (Pavia University and Salinas) demonstrate that classifiers trained on as few as 30 selected bands match or exceed full-spectrum baselines while reducing computational requirements. The resulting subsets align with physically meaningful, highly discriminative wavelength regions, indicating that model-aligned, explanation-guided band selection is a principled route to effective dimensionality reduction for HSI.
\end{abstract}

\keywords{Hyperspectral Image Analysis, Band Selection, Explainable Artificial Intelligence, LRP, SHAP, RISE.}

\section{Introduction}
\label{sec:intro}
 
{H}{yperspectral} imaging (HSI) is an advanced imaging technology that employs sensors to capture data across hundreds of narrow, contiguous spectral bands spanning a wide range of electromagnetic wavelengths well beyond the visible spectrum\cite{BHARGAVA2024e33208}.
Every pixel in a hyperspectral image records a complete light spectrum, yielding fine-grained spectral data that enables precise material identification and characterisation based on unique chemical and physical properties---capabilities that exceed those of conventional imaging systems. However, high-dimensionality presents significant challenges for analysis and modelling. It increases computational complexity, introduces redundancy, and exacerbates the risk of overfitting. Moreover, data quality is often compromised by noise from sensor imperfections, environmental factors, and external interferences, which can obscure meaningful spectral patterns.

To address these issues, researchers have widely adopted dimensionality reduction techniques~\cite{ALALIMI2023109096, rs14184579} aimed at eliminating redundancy, suppressing noise and retaining the the most informative spectral features. These techniques are broadly categorised into feature extraction and feature (band) selection, with different trade-offs in interpretability, efficiency, and downstream performance. 

Feature extraction~\cite{9082155, 10.1007/978-3-030-52190-5_12} transforms the original high-dimensional data into a lower-dimensional space where new representative feature vectors are selected but can obscure the physical and chemical significance of spectral bands---a critical limitation for applications that require domain-specific validation. 

In contrast, band selection~\cite{Goud2024, Sawant2021}, identifies and retains a subset of the original bands without altering their structure, thereby preserving their physical interpretability. 
Nonetheless, existing band-selection methods may mistakenly prioritise noisy bands as informative or capture specific spectral clusters, overlooking the broader spectral variability inherent in hyperspectral data. 
\begin{figure*}[t]
\centering
\includegraphics[width=0.95\textwidth]{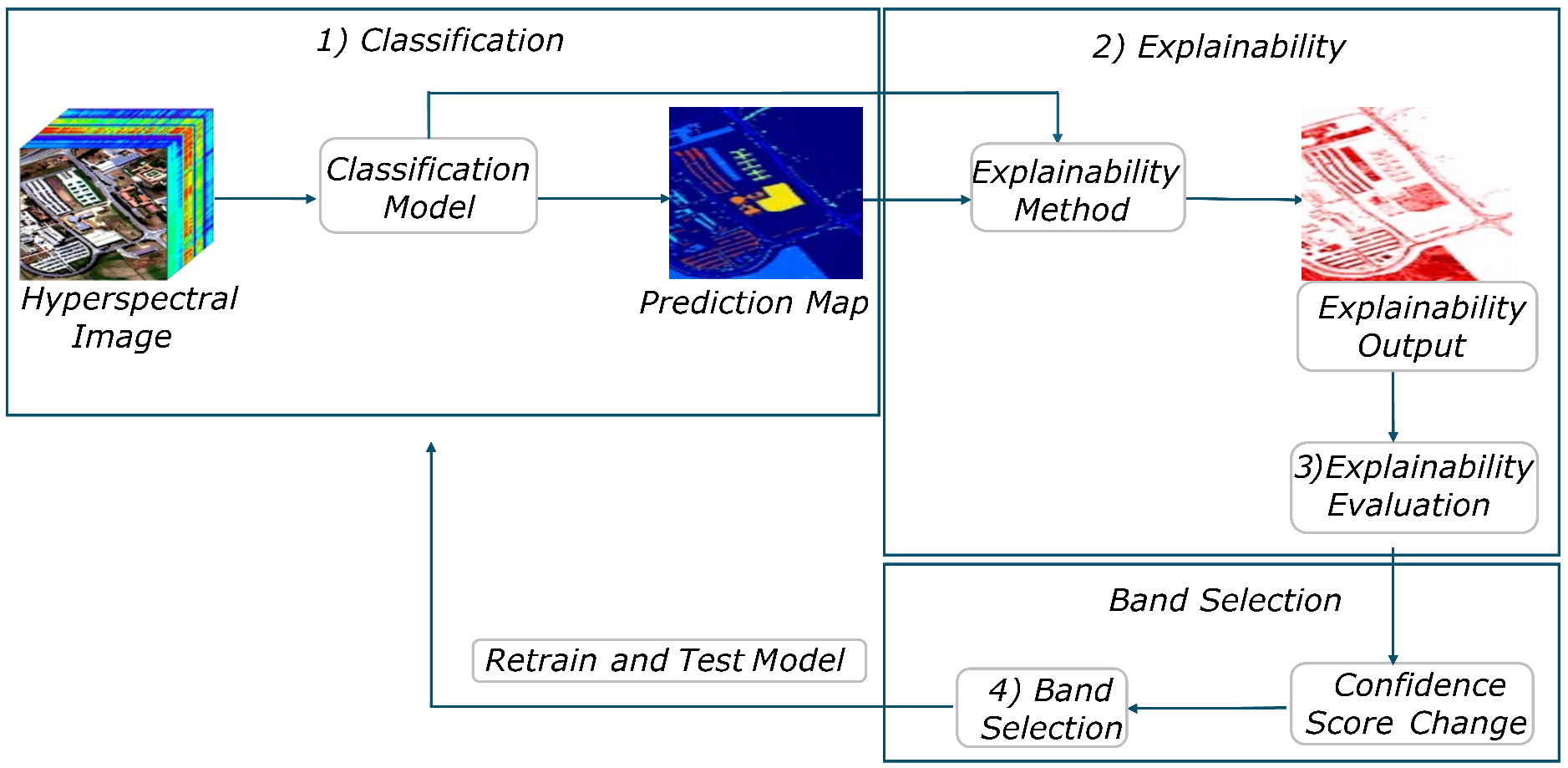}
 \caption {Workflow using explainability-based selection of key spectral bands and model retraining with reduced dimensionality.}
\label{fig1:method}
\end{figure*}

Meanwhile, there has been a growing emphasis on understanding models decision-making processes~\cite{doshi2017, ALI2023101805, https://doi.org/10.1002/widm.1493}, aiming to make complex models more transparent and interpretable. Explaining why a model makes a specific decision has driven research into model explainability~\cite{10.1145/2939672.2939778,LONGO2024102301,BARREDOARRIETA202082}. While explainability methods have been widely adopted to justify predictions, their integration into HSI analysis remains limited~\cite{10430776,10282341}. Beyond explaining model outputs, these methods have the potential to improve feature selection by identifying bands that significantly influence model predictions.
This explainability-driven band selection enables dimensionality reduction while maintaining robust performance. From an industrial perspective, isolating the most discriminative bands supports real-time applications by enabling faster image acquisition and enhancing overall efficiency.  

This paper is \emph{exploratory}: we examine whether post-hoc explanation methods (LRP, SHAP, and RISE) can guide hyperspectral band selection. For that we propose a four-stage methodology where we integrate explainability methods with Convolutional Neural Networks (CNNs) classifiers and prioritises influential spectral bands in line with model behaviour, avoiding statistical or heuristic criteria. This framework offers a promising direction towards advancing hyperspectral image analysis and presents the following key contributions are:

\begin{enumerate}[label=(\arabic*)]

\item \textbf{Exploratory evaluation of explainability for band selection:}
We conduct one of the first comparative studies of three established post-hoc explainability methods (LRP, SHAP, RISE) for band selection in HSI, assessed on two CNNs(TRI-CNN and HSI-CNN) and two benchmarks( PaviaU and Salinas),addressing a methodological gap in the literature.
 
\item \textbf{Physically faithful band selection:}
Band relevance is computed in the native wavelength domain, preserving the original band--wavelength correspondence and requiring no feature--space transformations or architectural changes.

\item \textbf{Accuracy-preserving dimensionality reduction:}
Classifiers using as few as \(30\) explainability-selected bands match or exceed full-spectrum baselines.

\end{enumerate}

 We organise the remainder of the paper as follows: Section~\ref{sec:related_work} reviews related work and positions our research with respect to existing efforts. Section~\ref{sec:material_method} describes the proposed method. In Section~\ref{sec:experiments_discussions}, we evaluate the considered explanation methods and empirically validate the explainability-based bands on model performance. Finally, Section~\ref{sec:conclusion} offers concluding remarks and outlines future work prospects.

\section{Related Work}
\label{sec:related_work}
 \noindent We  position our work along two main dimensions:

\textbf{\textit{Band selection methods.}} Dimensionality reduction constitutes a key preprocessing step in hyperspectral image analysis, with band selection often preferred due to its ability to retain original spectral information while reducing redundancy and computational complexity. Broadly, band-selection methods fall into three categories, search-based, learning-driven, and hybrid, each with distinct mechanisms, benefits and trade-offs.

\textit{Search-based methods}. These include \begin{enumerate*}[label=(\arabic*)]
\item Ranking, wherein each spectral band is assigned a score based on a predefined criterion, such as mutual information~\cite{Sawant2020ASO}, and the top scoring bands are selected. Although straightforward, ranking can be computationally expensive and may overlook inter-band dependencies, resulting in the selection of redundant bands. \item Clustering groups spectrally similar bands and selects representative bands from each cluster~\cite{YANG2017396}. While effective in reducing redundancy, performance of these methods relies heavily on the choice of clustering algorithm and its parameters. \item Heuristic search, such as genetic algorithms, iteratively refines subsets of bands to maximise performance~\cite{10037182, DEEP2024105053}, albeit with significant computational demands. \end{enumerate*}\\

\textit{Learning-driven methods}. These approaches integrate band selection directly into model training, allowing the model to highlight the most relevant bands~\cite{rs15184460}. \begin{enumerate*}[label=(\arabic*)]
\item  Embedded regularisation applies sparsity constraints to the model's input weights, ablating uninformative bands and flagging the remainder as important~\cite{8747509, 10089852}. However, these selections may not generalise across models.\item Deep learning, such as autoencoders, compress and reconstruct the full spectrum; analysing reconstruction errors or bottleneck activations reveals the most informative  bands~\cite{9032380}. Alternatively, attention mechanisms in CNNs or transformers learn per-band weights during training, selecting those with consistently high scores~\cite{8972909, Kumar2022AttnAE}. 
Both mechanisms effectively learn compact and performant spectral representations, yet they introduce computational overhead.  
\end{enumerate*}\\

\textit{Hybrid methods}. These methods combine elements from search-based and learning-driven strategies aiming to balance computational cots and selection accuracy~\cite{9817391,zimmer2024embeddedhyperspectralbandselection}. 

In contrast to conventional approaches, we harness explainability methods to identify spectral bands that strongly drive predictions. This allows us to avoid the drawbacks of high computational overhead and inter-band correlations typically associated with conventional band selection.

\textbf{\textit{Model Explainability}}. We focus on the post-hoc explainability methods that justify the decisions of a pre-trained model. In image analysis, these methods assign importance to pixels or regions, highlighting features that drive model predictions. In the hyperspectral imagery, the spectral dimension maps each input channel to a specific wavelength, enabling band-level attributions. For instance,~\cite{10430776} applies the model agnostic SHAP algorithm to a CNN-based hyperspectral remote sensing image classifier to explain model outputs. A preprocessing step using Principal Component Analysis (PCA)~\cite{doi:10.1080/02564602.2020.1740615} is applied to reduce dimensionality by projecting the original pixel-wise band values onto a new set of orthogonal components. This step removes the direct band--wavelength correspondence, thus preventing us from attributing the model decisions back to the individual bands and undermining spectral interpretability. Similarly,~\cite{10282341} employs back-propagation-based methods such as GradCAM~\cite{selvaraju2020gradcam}, GradCAM++~\cite{8354201} and Guided Back-propagation~\cite{springenberg2015strivingsimplicityconvolutionalnet} to highlight critical spectral regions, but reports mixed results regarding band-specific importance. Their results indicate that while Guided Back-propagation identifies critical spectral bands for certain classes, the overall band importance remained relatively uniform, underscoring the need for sharper wavelength-level attributions. While~\cite{ZHANG2023100491} integrates multiple explainability methods for band selection, it primarily evaluates combinations of CNN architectures and normalisation strategies. Although it reports performance with selected bands, it lacks a direct comparison of the relative effectiveness of the explainability methods used, their contributions, quality or suitability for guiding the selection.

Our research leverages explainability to guide band selection in HSI. By identifying and validating the bands that have the highest influence on the model's decisions, our approach simultaneously reduces dimensionality, maintains interpretability, and enhances computational efficiency. 

\section{Model design and description}
\label{sec:material_method}
In this section, we outline our methodology for band selection using explainability techniques. Figure~\ref{fig1:method} provides an overview of the four stages involved in the process. 

\subsubsection*{Stage \(1\): Classification}
 \label{subsubsec:stage1-classification}

We begin by densely extracting overlapping patches from the hyperspectral image \(X\in \mathbb{R}^{h\times w\times b}\) where \(h\) and \(w\) are the spatial dimensions and \(b\) the number of spectral bands. Each patch \(X_p\in \mathbb{R}^{h'\times w'\times b}\) corresponds to a small spatial region of \(X\) of size \(h'\) and \(w'\), retains all \(b\) bands, and is centred on  pixel \((i,j)\) from which it inherits its ground-truth label. We define a classifier as the mapping
\begin{equation}
\label{eq:classifier}
F:\mathbb{R}^{h'\times w'\times b}\;\longrightarrow\;\mathbb{R}^c,
\end{equation}
where \(c\) is the number of classes. For each input patch \(X_p\), 
the classifier produces a confidence (logit or probability) vector
\begin{equation}
\label{eq:classifier_output}
F(X_p) = \bigl(F(X_p)^{(1)},\ldots,F(X_p)^{(c)}\bigr)\in \mathbb{R}^c
\end{equation}
We obtain the predicted class by the maximum confidence rule:
\begin{equation}
\label{model_prediction}
Y_{p} = \arg\max_{k \in \{1, \dots, c\}} F(X_{p})^{(k)}.
\end{equation}

\subsubsection*{Stage \(2\): Explainability}
\label{subsec:stage2}
In the second stage, an explainability method \(E\) is applied to analyse the 
classifier and its decision in order to assign importance values to each spectral band. We formalise it as
\begin{equation}
\label{eq:explainability}
E:\;\mathcal{F}\times\mathbb{R}^{h'\times w'\times b}\times\mathcal{R} 
   \;\longrightarrow\;\mathbb{R}^b 
\end{equation}
where \(\mathcal{F}\) denotes the space of classifiers, 
\(\mathbb{R}^{h'\times w'\times b}\) is the input patch space, 
and \(\mathcal{R}\) represents the internal model representations 
(e.g.\ feature maps or latent activations). 
The output \(O = E(F, X_p, R)\in\mathbb{R}^b\) is a band-relevance vector 
that assigns an importance score to each spectral band for the model’s decision on patch \(X_p\).

\subsubsection*{Stage \(3\): Evaluation}
\label{subsec:stage3}
We assess each explanation method by systematically measuring how the model’s 
confidence changes as bands are progressively perturbed according to their 
relevance scores. Specifically, bands are either \emph{deleted} (replaced by the 
training-set mean) or \emph{inserted} (restoring the original values). At each step, we modify \(20\%\) of the bands, in descending order of relevance, and record the resulting change in confidence. Formally, we define the evaluation operator
\begin{equation}
\label{eq:evaluation}
V:\;\mathbb{R}^b_{\geq 0}\times \mathbb{R}^{h'\times w'\times b}\times \mathcal{F}
   \;\longrightarrow\;\mathbb{R},
\end{equation}
which, when applied to a relevance vector \(O\in\mathbb{R}^b_{\geq 0}\), 
a patch \(X_p\in\mathbb{R}^{h'\times w'\times b}\), and a classifier 
\(F\in\mathcal{F}\), yields
\begin{equation}
Q \;=\; V(O, X_p, F) \in \mathbb{R},
\end{equation}
where \(Q\) is the evaluation score that quantifies the confidence drop 
(for deletion) or confidence gain (for insertion). This procedure reveals 
how faithfully the relevance vector \(O\) captures the model’s reliance on 
individual bands.

\subsubsection*{Stage \(4\): Band Selection and Retraining}
\label{subsec:stage4}
Based on the evaluation results, bands whose removal elicit the largest confidence drop or whose insertion yields the largest gain, are deemed most relevant. We aggregate and normalise these per-band confidence changes into a single influence score and select the top \(b'\) bands.
Formally, let 
\( \mathcal{B}' \subset \{1,\dots,b\}, \quad |\mathcal{B}'| = b'\) denote the selected subset of bands, where \( b' < b \). For each original patch \(X_p\), we construct \(X'_p \in \mathbb{R}^{h'\times w'\times b'}\), by restricting \(X_p\) to the bands in \(\mathcal{B}'\). We then retrain an adjusted classifier \(F':\;\mathbb{R}^{h'\times w'\times b'} \;\longrightarrow\; \mathbb{R}^c\), so that for each reduced patch \(X'_p\) the prediction is \(Y'_p = F'(X'_p).\)

\section{Experiments}
\label{sec:experiments_discussions}
 \subsection{Data}
\label{subsec:data}
We validate our methodology on two publicly available hyperspectral benchmarks: Pavia University (PaviaU) and Salinas~\cite{ehu_hyperspectral}. PaviaU, acquired by the ROSIS sensor, comprises 103 spectral bands spanning \SI{0.43}{\micro\metre} to \SI{0.86}{\micro\metre} at \SI{1.3}{\metre} per--pixel spatial resolution, with image dimensions of $610 \times 340$ pixels and 9 ground-truth classes.
Salinas, captured by the AVIRIS sensor, originally contains \(224\) spectral bands spanning \SI{0.4}{\micro\metre} to \SI{2.5}{\micro\metre}; after removing water-absorption bands, 204 bands remain. Its spatial resolution is \SI{3.7}{\meter} per pixel, and has $512\times 217$ pixels. It encompasses \(16\) ground-truth classes.

For both datasets, we apply the same dense, overlapping patch extraction described in section~\ref{sec:material_method}, Stage \(1\): Classification. This yields approximately \(42,318\) labelled patches for PaviaU and \(54,129\) for Salinas, each patch inheriting the class of its centre pixel.
\subsection{Neural Network Architecture}
\label{subsec:nn_arch}
To validate our approach, we consider two CNN architectures for hyperspectral remote sensing classification, using their predictions as baseline for evaluation. These architectures are not intended for benchmarking or state-of-the-art performance; rather, they serve as test beds to validate the utility of explainability methods in guiding band selection in our experiments.\\
\noindent\textit{Tri-CNN}~\cite{rs15020316}. A multi-scale 3D-CNN with three parallel branches (spectral, spatial, and spectral-spatial). Feature maps of each branch are flattened, concatenated, and fed to fully connected layers and a softmax layer. We use \(13 \times 13 \times b\) (PaviaU) and \(11 \times11 \times b\) (Salinas) patches retaining the full spectral dimension \(b\), unlike the original work which reduces the spectral dimension to \(15\) and \(35\) components, respectively, using PCA. For PaviaU, we follow the original split of \(1\%\) train, \(99\%\) test. For Salinas, due to its higher spectral dimensionality, we use \(75\%\) train, \(25\%\) test, to manage the computational load.\\
\noindent\textit{HSI-CNN}~\cite{8455251}. A two-dimensional (2D) CNN that converts one-dimensional spectral vectors into 2D matrices and processes them by standard 2D convolutions to extract spectral-spatial features. We use \(3 \times 3 \times b\) patches on both datasets and the original split of \(30\%\) train/ \(70\%\) test.

Under both methods, we split each model's designated test set into two equal subsets. One is used for assessing the classification accuracy, and the other is reserved exclusively for generating and evaluating the explanation maps.

\begin{figure*}[ht]
\centering
\includegraphics[width=0.95\textwidth]{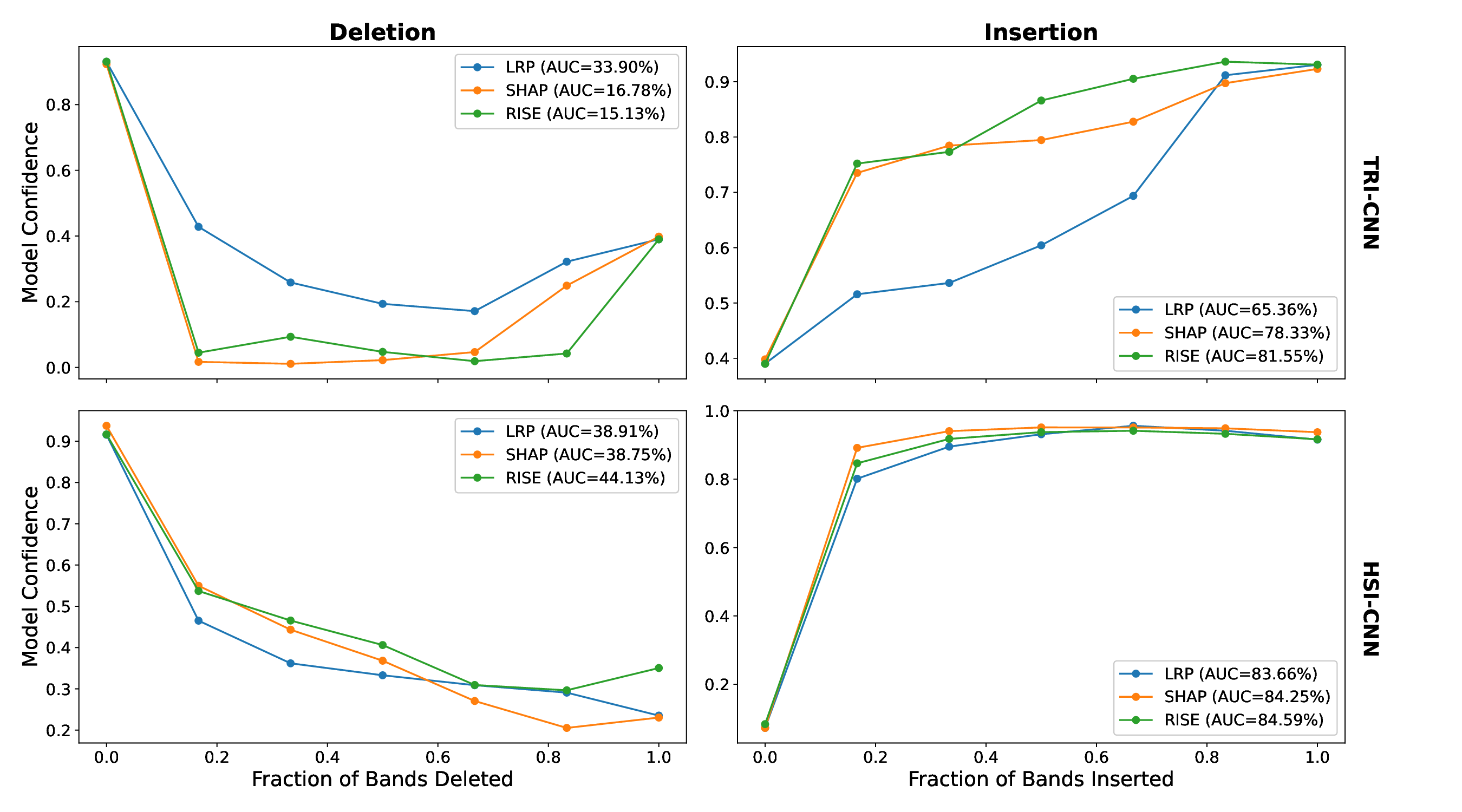}
  \caption{PaviaU: Deletion / Insertion AUC curves. TRI-CNN (top), HSI-CNN (bottom).}
  \label{fig2:paviau_del_insrt}
\end{figure*}

For Stage \(2\), we implement three explainability methods:\\

\noindent\textbf{Layer-wise Relevance propagation (LRP)}~\cite{10742949} attributes the neural network’s decision by redistributing its output backwards through each layer according to predefined rules described in~\cite{10.1371/journal.pone.0130140,Montavon2019}. We first perform a standard forward pass to store the activation maps. Relevance is then initialised at the networks' output layer using the logit of the true class and propagated backward. We apply different LRP rules at various layers: \begin{enumerate*}[label=(\arabic*)]\item the \textbf{LRP-0} rule for fully connected layers, which distributes relevance proportionally to neuron activations;
\item the \textbf{$\epsilon$-rule} for intermediate layers, introduces a small stabiliser $\epsilon$ to suppress noise and avoid division by zero; \item the \textbf{$\gamma$-rule} for early layers, which scales positive weights by a factor $\gamma>0$ to emphasise positive contributions (set to 0.25 in our experiments).\end{enumerate*} 
The output is a relevance map \( O_{\mathrm{LRP}} \in \mathbb{R}^{h' \times w' \times b} \) where each value reflects the contribution of a spectral band and spatial location to the model's decision. 

\noindent\textbf{SHAP (SHapley Additive exPlanations)}~\cite{NIPS2017_8a20a862} assigns a Shapley value to each input feature, quantifying its contribution to the model's output. 
The values are grounded in cooperative game theory and are computed by evaluating every possible feature subset. However, this exhaustive computation is infeasible for high-dimensional data.
To balance accuracy and efficiency, we sample \(30\) random subsets of bands. For each subset, we compute two predictions: one including the band, \(p_{\mathrm{incl}}=F(X_p^{S \cup \{\text{band}\}})\), and one excluding it, \(p_{\mathrm{excl}}=F(X_p^{S})\).
The band’s marginal contribution for that subset is computed as $(p_{\mathrm{incl}}{-}p_{\mathrm{excl}}$). Averaging these differences over the \(30\) subsets yields an approximate Shapley value for each band. Repeating this for all \( X_p \) patches produces a band-level importance matrix \(O_\text{SHAP} {\in} \mathbb{R}^{X_p {\times} b} \).

\noindent\textbf{Randomized Input Sampling for Explanation (RISE)}~\cite{Petsiuk2018RISERI} explains the model output using perturbations to generate saliency maps. Unlike gradient-based methods, RISE is architecture-agnostic and requires no access to the internal parameters of the model. In our implementation, we sample (\(5000\)) one-dimensional spectral masks that randomly occlude different bands while preserving the patch spatial layout. We apply a \textbf{density} ratio of \(0.5\), indicating that \(50\%\) of the input remains visible in each mask. Because masks are constant over the spatial extent, the procedure yields \emph{band-level} (spatially uniform) relevance scores for each instance. Hence RISE returns a \emph{band-level} relevance vector \(O_{\mathrm{RISE}}(x)\in\mathbb{R}^{b},\) i.e., one score per band. Stacking across \(X_p\) patches \(\{x_i\}_{i=1}^{X_p}\) gives \(O_{\mathrm{RISE}}\in\mathbb{R}^{X_p\times b}.\)

\subsection{Explainability Assessment Framework}
\label{subsec:explanability_assessment_fw}
We evaluate the faithfulness of the explanation-maps using two perturbation-based metrics~\cite{Petsiuk2018RISERI}, which quantify how changes to the most relevant features impact the prediction.

\textbf{\textit{Deletion -- Insertion}}~\cite{wang2024benchmarking}: measures how the model's confidence changes as we gradually remove or add the most relevant bands.
    
\noindent{\emph{Ranking bands}}. For a given input, we sum each band's positive relevance scores over the spatial dimension, this gives a single relevance value per band. We then sort bands from the most to the least relevant.

\noindent{\emph{Deletion}}. Starting with the intact input, we replace bands in \(20\%\) increments with their training-set mean values in the ranked order, continuing until all bands are replaced. After each increment, we record the confidence in the true class; a steeper decline indicates that the explanation method has correctly identified bands critical to the model's decision.

\noindent{\emph{Insertion}}. Starting from an empty input (all bands are set to zero), we reintroduce band values in descending order of relevance, progressively in \(20\%\) increments, tracking the increase in the confidence gain. A steeper rise in confidence shows that early-added bands drive prediction, indicating a more faithful explanation.  

We summarise each process by the area under its confidence–fraction curve (AUC): lower AUC is better for deletion (downward trend), and higher AUC is better for insertion (upward trend). Together, these AUC values quantify how effectively an explanation method identifies the most important bands.

\textbf{\textit{Average--Drop}}. This metric measures how much the model’s confidence in the true class falls when low‐relevance bands are masked out. Specifically, it measures whether an explanation map effectively identifies critical features. Concretely, we perform an element-wise multiplication between the original input and its relevance map, preserving the most influential bands while suppressing the weakly relevant ones. We then feed the result through the model and measure the decrease in the model's confidence for the true class. The percentage drop in class confidence reflects the overall quality of the explanation: smaller drops imply that the preserved bands capture the core discriminative information and thus a higher-quality explanations.
 
\begin{figure*}[ht]
  \centering
 
\includegraphics[width=0.95\textwidth]{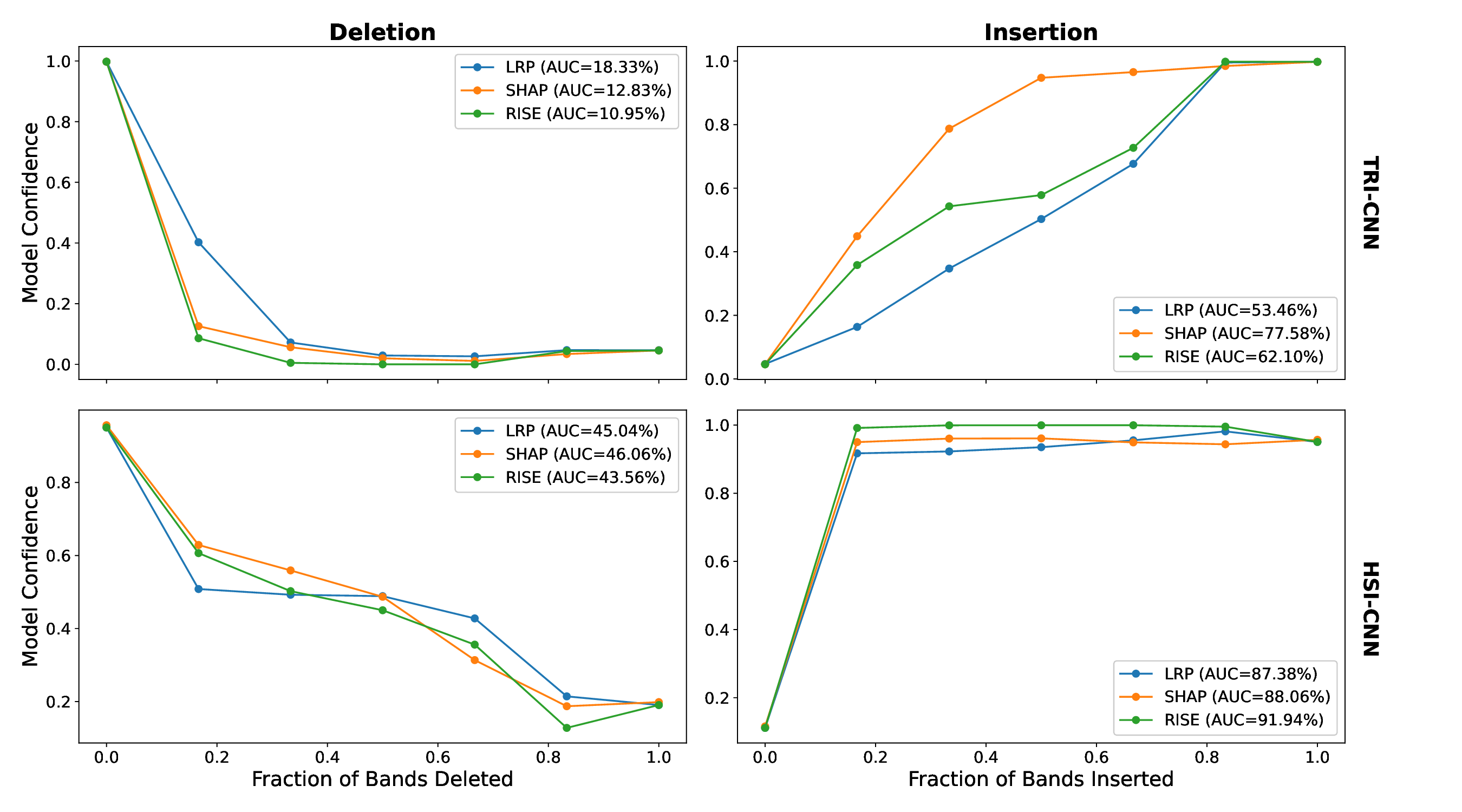}
  \caption{Salinas: Deletion / Insertion AUC curves. TRI-CNN (top), HSI-CNN (bottom).}
  \label{fig3:salinas_del_insrt}
\end{figure*}

\subsection{Explainability Evaluation Results}
\label{subsec:evaluation_results}
Figures~\ref{fig2:paviau_del_insrt},~\ref{fig3:salinas_del_insrt} display the AUC curves obtained from the \textit{{Deletion--Insertion}} tests. In each case, the confidence decreases under deletion and increases under insertion, exhibiting the expected trends. The magnitude and rate of these changes differ between the models, which we attribute to differences in their architectural complexity.
Replacing bands with their mean values rather than entirely removing them, prevents the confidence from reaching zero. Moreover, using mean values aligns more closely with the distribution learned by the model, causing a partial rebound in confidence after the replacement of the final fraction of bands.


 \begin{table}[t]
  \centering
   \small                           
  \setlength{\tabcolsep}{6.5pt}      
  \renewcommand{\arraystretch}{1.15}  
  \caption{Average Drop (\%) comparison across explainability methods for Pavia and Salinas.}
  \label{tab_1:avg_drop}
  \begin{tabular}{llccc}
    \toprule
    Model & Dataset & LRP (\%) & SHAP (\%) & RISE (\%) \\
    \midrule
    \multirow{2}{*}{TRI-CNN} & Pavia   &  2.006 &  3.286 & 12.381 \\
                             & Salinas & 26.568 &  7.082 & 15.022 \\
    \midrule
    \multirow{2}{*}{HSI-CNN} & Pavia   &  0.218 &  7.527 &  1.317 \\
                             & Salinas &  1.795 &  4.981 &  0.250 \\
    \bottomrule
  \end{tabular}
\end{table}

Table~\ref{tab_1:avg_drop} reports the \textit{\(Average\- Drop\)} (\(\%\)) in model confidence across the three explainability methods. It is critical to highlight that band-level explanations may lack the granularity provided by pixel-level analyses, potentially leading to the omission or under-representation of truly important features. This loss of detail can lead to a larger drop in model confidence. Nonetheless, this band-wise evolution remains effective in comparing the relative ability of the different methods to identify the most influential spectral information.
  
\subsection{Explainability-Driven Top-30 Band Selection}

\begin{table}[ht]
\centering
 \setlength{\tabcolsep}{6.5pt}      
\renewcommand{\arraystretch}{1.15}  
\caption{Average test accuracy ($\%$) using the full bands and a subset of 30 bands selected based on explainability methods for PaviaU and Salinas datasets.}
\label{tab_2:accuracy_results}

\begin{tabular}{llcccc}
\hline
Model    & Dataset  & Full Bands & LRP-30 & SHAP-30 & RISE-30 \\ \hline
\multirow{2}{*}{TRI-CNN}  & Pavia    & 91.94 ± 1.67 & 91.68 ± 0.56 & 91.47 ± 0.46 & 91.21 ± 0.65 \\  
                          & Salinas  & 99.85 ± 0.02 & 99.82 ± 0.09 & 99.86 ± 0.03 & 99.86 ± 0.03 \\ \hline
\multirow{2}{*}{HSI-CNN}  & Pavia    & 95.25 ± 0.16 & 92.92 ± 0.27 & 92.98 ± 0.26 & 92.94 ± 0.14 \\  
                          & Salinas & 97.30 ± 0.25& 94.89 ± 0.13 &94.75 ± 0.38 & 95.94 ± 0.25   \\ \hline
\end{tabular}

\end{table}
Building on the procedure in Section~\ref{subsec:stage4}---Stage 4, we employed the \textit{{Deletion -- Insertion}} evaluation method to identify the \(30\) bands with the highest influence on model confidence.
Each network was then retrained and evaluated over five independent runs using the same hyperparameters as in the original training on the full spectral bands, adjusting only the kernel sizes and strides to accommodate the reduced spectral dimensionality.

Table~\ref{tab_2:accuracy_results} compares the test accuracies of TRI-CNN and HSI-CNN trained on the full set of spectral bands versus the \(30\) bands selected using the explainability methods. Despite the reduced spectral dimensionality, TRI-CNN maintains nearly identical performance across all band-selection strategies on both datasets, suggesting it can effectively adapt to fewer bands without a significant loss in accuracy. By contrast, the HSI-CNN, which relies more heavily on the full spectral range, exhibits a noticeable decline in performance, albeit to varying extents. On PaviaU, accuracy drops by approximately \(2.27\%\)--\(2.33\%\), and on Salinas \(1.36\%\)--\(2.55\%\), indicating a higher sensitivity to spectral reduction.
\begin{figure*}[th!]
\centering
\includegraphics[width=0.95\textwidth]{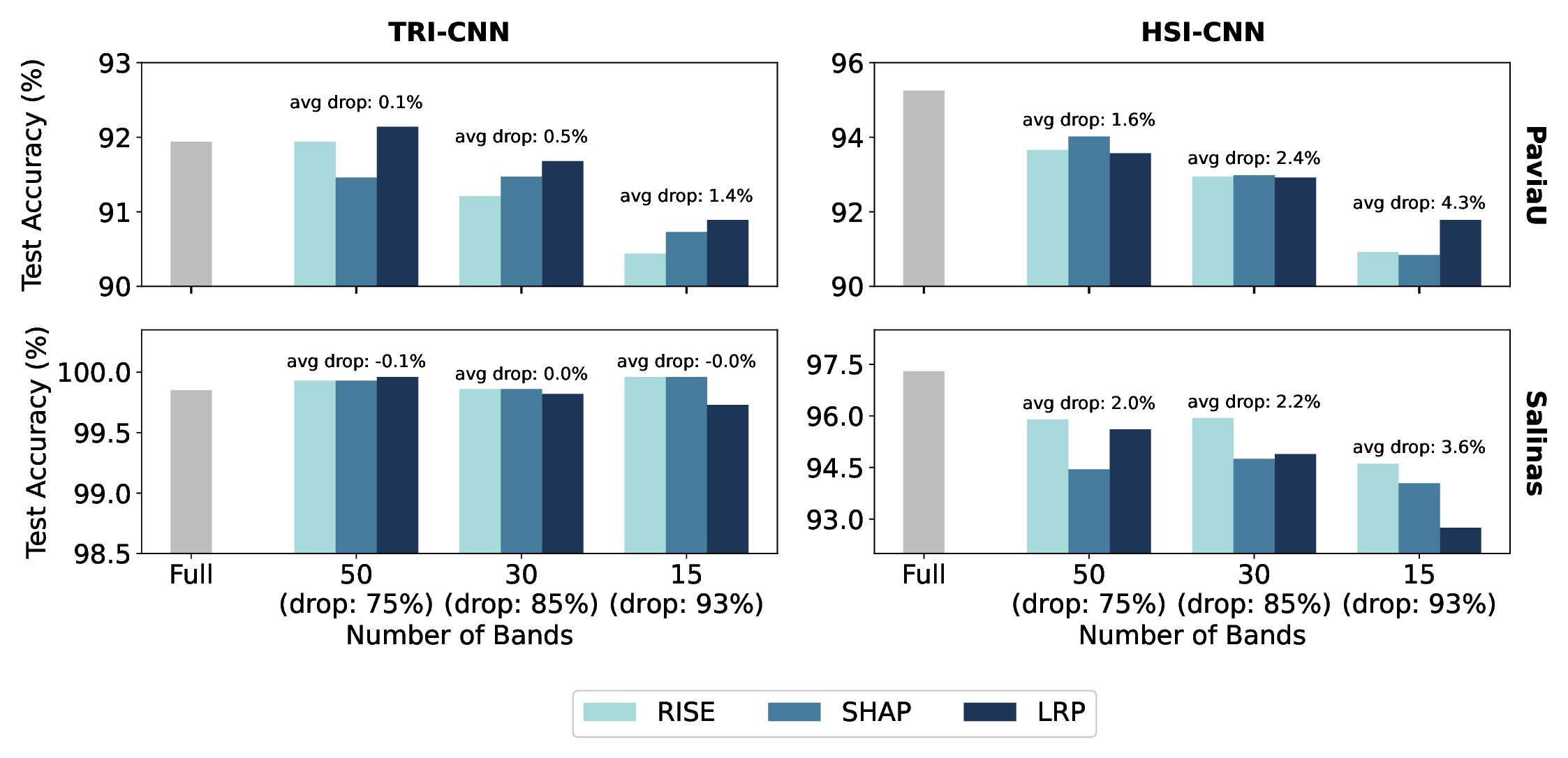} %
\caption {Test Accuracy Performance (\%) of TRI-CNN (left) and HSI-CNN (right) with Different Subsets of Selected Bands (Full, 50,30,15). PaviaU (top), Salinas (bottom).}
\label{fig4:bands_subsets}
\end{figure*}
Given the intricate correlations among spectral bands, one might expect an even larger accuracy drop. However, our explainability-driven selection mitigates this effect by preserving the most informative bands, thereby limiting the loss of critical information. Overall, the results demonstrate that our explainability-based approach achieves effective dimensionality reduction while preserving robust performance.

\subsection{Effect of the Number of Selected Bands}

To evaluate the suitability of the \(30\)-band subset used previously, we compare model performance with \(15\)- and \(50\)-band subsets for both datasets (Figure~\ref{fig4:bands_subsets}).
As expected, both models exhibited accuracy losses compared to using the full spectral range. However, the explainability-driven band-selection successfully isolated the most informative bands, keeping high accuracy even with only \(15\) bands. On PaviaU, both models maintained over \(90\%\) accuracy at \(15\) bands, while on Salinas the TRI-CNN remained near \(100\%\) and the HSI-CNN above \(94\%\).

When we increased the bands to \(50\), both models improved on both datasets, yet with diminishing returns beyond \(30\) bands. 
For instance, on PaviaU, TRI-CNN's accuracy rose approximately \(0.7\%\)--\(1.5\%\) across RISE, SHAP, and LRP when going from \(15\) to \(30\) bands, yet only yielded another \(0.7\%\) when bands were increased to \(50\). A similar trend appeared on Salinas, where the TRI-CNN’s accuracy change after \(30\) bands was negligible (\(\leq 0.1\%\)), highlighting its robustness to spectral-band reduction. The HSI-CNN showed larger overall gains, \(1.8\%\)--\(3.2\%\) for PaviaU and \(2.0\%\)--\(3.6\%\) for Salinas from \(15\) to \(50\) bands, but again we saw only marginal gains (\(0.65\%\)–-\(1.0\%\)) beyond \(30\) bands.

Thus, while adding bands can boost accuracy, benefits plateau after \(30\) bands, suggesting that a \(30\)-band subset strikes an optimal balance between computational efficiency and classification performance for both datasets.

\subsection{Distribution of Selected Bands' Wavelengths} 

\begin{figure*}[t]
  \centering
    \includegraphics[width=0.95\textwidth, height=9cm]{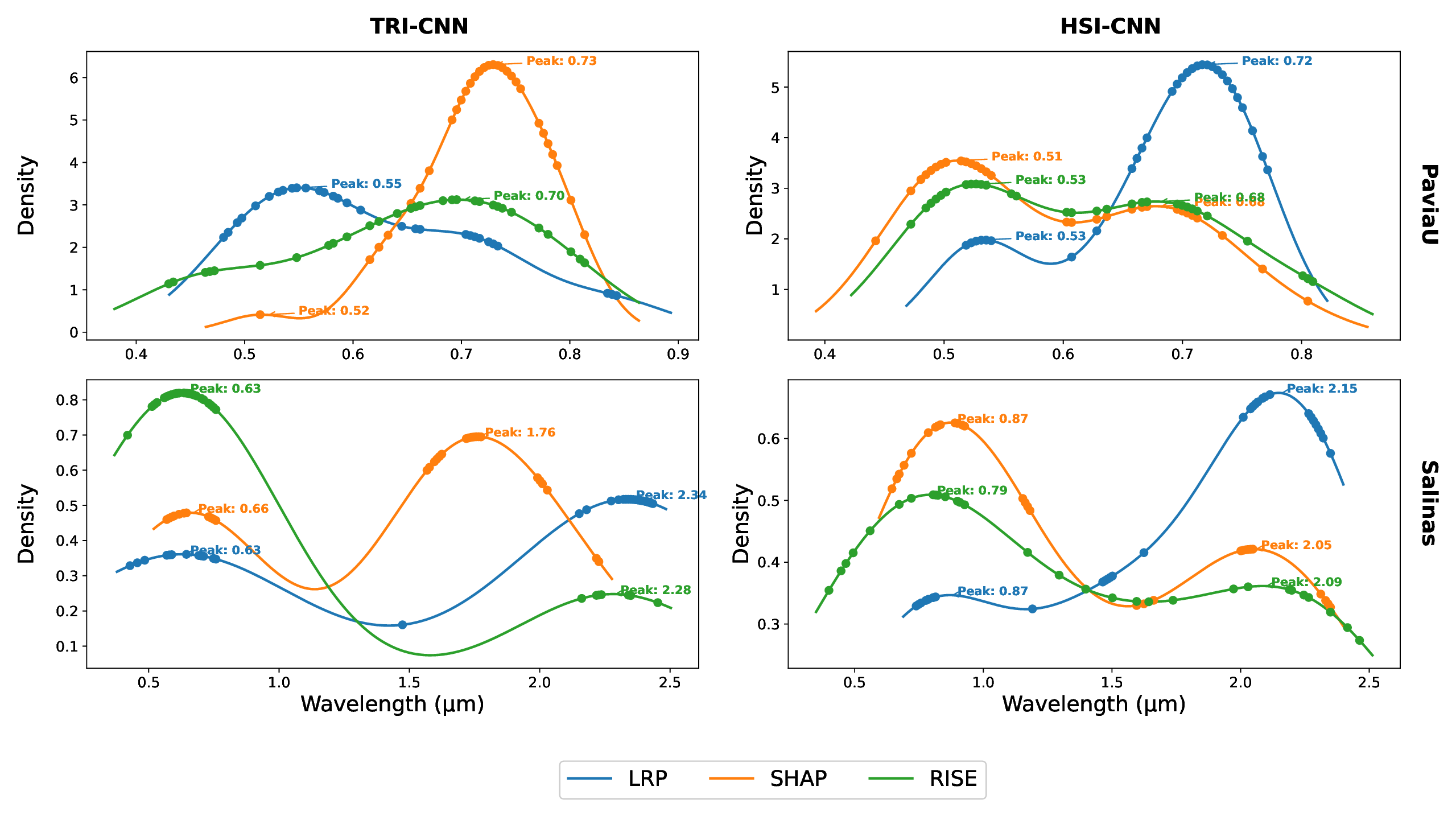}%
  \caption{Distribution of selected bands wavelength-ranges (in \(\mu\)m) by each explanation method for TRI-CNN~(left), HSI-CNN~(right), PaviaU (top), Salinas (bottom).}
  \label{fig5:wl_ranges_kde}
\end{figure*}

To investigate whether the different explainability methods converge on similar spectral regions (i.e.\ whether they follow similar wavelength distributions), we performed a kernel density estimation (KDE) over the wavelengths of the top \(30\) bands selected by each method (Figure~\ref{fig5:wl_ranges_kde}). 
We computed KDEs using \texttt{gaussian\_kde} with Scott's rule \begin{math} h = \sigma \, n^{-\frac{1}{d + 4}}\end{math} which automatically determines a data-dependent bandwidth based on the spread of values within each subset. As a result, each curve may have a slightly different bandwidth, reflecting variations in the selected wavelength distributions across methods.

\paragraph{\textbf{Pavia University (Urban scene).}} 
Although each explanation method selects slightly different spectral regions, all three method consistently concentrate their selections within the visible--red-edge interval  (\SIrange{0.43}{0.84}{\micro\metre}). According to~\cite{https://doi.org/10.1029/2024JH000464}, this region coincides with the chlorophyll absorption peak (\SI{0.67}{\micro\metre}) and the red-edge (\SIrange{0.68}{0.75}{\micro\metre}), which underpin vegetation indicators and help differentiate vegetated (natural) from man-made (built) surfaces. Bands in this interval are highly discriminative for urban land-cover analysis (e.g.\, asphalt, concrete, roofing, and vegetation).
    
For TRI-CNN, the KDE peaks occur at \SI{0.55}{\micro\metre} (LRP), \SI{0.73}{\micro\metre} (SHAP), and \SI{0.70}{\micro\metre} (RISE), with all three distributions concentrated in the mid-visible region. Although LRP's selections extend numerically to \SI{0.84}{\micro\metre}, its primary focus lies between \SIrange{0.48}{0.76}{\micro\metre}.

For HSI-CNN, dual peaks appear near \SI{0.53}{\micro\metre} and \SI{0.72}{\micro\metre} for LRP, \SI{0.51}{\micro\metre} and \SI{0.68}{\micro\metre} for SHAP, and \SI{0.53}{\micro\metre}and \SI{0.68}{\micro\metre} for RISE. These broader selections (\SIrange{0.44}{0.81}{\micro\metre}) still emphasise the mid-visible region, despite their wider span of the spectrum.
\paragraph{\textbf{Salinas dataset (Agricultural Scene).}} 
All three methods show prominent selections in the visible/red edge range around \SIrange{0.5}{0.75}{\micro\metre}and the short‐wave infrared (SWIR) region \SIrange{1}{2.15}{\micro\metre}. The visible--red-edge bands capture key pigment and structural variations in vegetation canopies, while the SWIR bands probe a range critical for characterising vegetation health, moisture content levels, and soil properties. According to~\cite{https://doi.org/10.1029/2024JH000464,VANDERMEER200455}, SWIR wavelengths penetrate deeper into canopy layers and are sensitive to dry matter and leaf internal structure, making them particularly informative for agricultural modelling. TRI-CNN's KDE nearly covers the entire visible--NIR range, while HSI-CNN shows two distinct peaks, one in the high visible--low-NIR region,\SIrange{0.79}{0.87}{\micro\metre}, and another in the SWIR (\SIrange{2.00}{2.15}{\micro\metre}).
 
Although each explanation method selects slightly different spectral regions, they consistently highlight overlapping bands. For PaviaU, all methods converge on mid--visible wavelength (\SIrange{0.43}{0.84}{\micro\metre}), while for Salinas, the selections span the visible into NIR, underscoring the importance of both spectral regions.~These patterns indicate a dual alignment: the selected bands are \emph{model-aligned}, reflecting the classifiers’ decision process, and \emph{domain-meaningful}, coinciding with established spectral features (e.g.\ chlorophyll absorption, SWIR sensitivities); supporting explainability-driven band selection as a principled route to compact, interpretable spectral subsets.

\section{Conclusion}
\label{sec:conclusion}
We demonstrate that post-hoc explainability methods can effectively drive dimensionality reduction in hyperspectral imaging by identifying the most informative spectral bands and wavelength ranges. Despite their differing complexities, both CNN architectures maintained robust classification performance when retrained on the reduced spectral subset. Furthermore, all three methods consistently converged on similar wavelength regions, highlighting the robustness and physical relevance of the selected bands. Unlike conventional band selection techniques, which risk discarding critical signals, overlooking data diversity, or erroneously prioritising noise, our approach preserves the chemical and spectral integrity of the data while reducing redundancy. In future work, we will validate these findings on additional datasets and evaluate other families of explainability methods to further generalise our results.\\

 
\section*{Acknowledgments}
The research presented in this article is part of the project "Learning-based representations for the automation of hyperspectral microscopic imaging and predictive maintenance "funded by the Flanders Innovation \& Entrepreneurship--VLAIO, under grant number HBC.2020.2266.

\bibliographystyle{unsrt}  
\bibliography{references}

\end{document}